\documentclass{article} 
\usepackage{iclr2026_conference,times}

\usepackage{amsmath,amsfonts,bm}

\def\eqref#1{equation~\ref{#1}}

\def\1{\bm{1}}

\DeclareMathAlphabet{\mathsfit}{\encodingdefault}{\sfdefault}{m}{sl}
\SetMathAlphabet{\mathsfit}{bold}{\encodingdefault}{\sfdefault}{bx}{n}

\usepackage{microtype}
\usepackage{hyperref}
\usepackage{url}
\usepackage{booktabs}
\usepackage{amsmath}
\usepackage{lineno}
\usepackage{graphicx}
\usepackage{subfigure}
\usepackage{wrapfig}
\usepackage{xspace}
\usepackage[normalem]{ulem}
\usepackage{enumitem}
\usepackage{multirow}
\usepackage{hyperref}
\usepackage{url}
\usepackage{csquotes}
\usepackage[font=small]{caption}
\usepackage{tcolorbox}
\usepackage{xcolor}      
\usepackage{colortbl}    
\usepackage{booktabs}    
\usepackage{fontawesome5}

\title{Pushing Test-Time Scaling Limits of \\ Deep Search with Asymmetric Verification}

\author{Weihao Zeng$^1$ \quad Keqing He$^{2}$\quad Chuqiao Kuang$^3$\quad Xiaoguang Li$^3$\quad Junxian He$^1$  \\
$^1$The Hong Kong University of Science and Technology \quad
$^2$BUPT
\quad 
$^3$Huawei Noah's Ark Lab \\
\texttt{\{wzengak,junxianh\}@cse.ust.hk} \quad \faGithub \ \href{https://github.com/hkust-nlp/deepsearch-tts}{hkust-nlp/deepsearch-tts}
}

\iclrfinalcopy 
\begin{document}

\maketitle



\begin{abstract}
Test-time compute can be scaled both sequentially and in parallel. Sequential scaling involves lengthening the generation process, while parallel scaling involves verifying and selecting among multiple candidate outputs. Combining these two strategies has led to the most powerful AI systems, such as Grok 4 Heavy and GPT-5 Pro. In certain contexts (e.g., solving Sudoku puzzles), verifying responses can be substantially easier than generating them. This property, referred to as \emph{asymmetric verification}, highlights the strong potential of test-time scaling (TTS).
In this work, we study both sequential and parallel TTS of deep search agents, motivated by the intuition that verification in this setting is often much easier than generation. 
In experiments, we first show that sequential scaling methods, such as budget forcing, can be effective initially but soon degrade performance. Leveraging asymmetric verification, however, we are able to achieve substantial improvements by allocating only a modest amount of compute to the verifier. 
We conduct experiments with flagship open-source models
and extend them to their ``Heavy'' variants through TTS. These deep research agents achieve gains of up to 27 absolute points on benchmarks such as BrowseComp. Remarkably, as an open-source alternative, GLM-4.5 Heavy reaches accuracy of {\bf 54.0\%} on BrowseComp and {\bf 66.0\%} on GAIA,  
placing it comparable to the best proprietary choices such as OpenAI Deep Research. Tongyi-DeepResearch Heavy further achieves {\bf 69.0\%} accuracy on BrowseComp, 
greatly surpassing the best proprietary results.

\end{abstract}

\begin{figure}[h]
\vspace{-15pt}
        \centering
    \begin{minipage}{0.96\textwidth}
    \centering
        \subfigure{
                \includegraphics[width=0.48\textwidth]{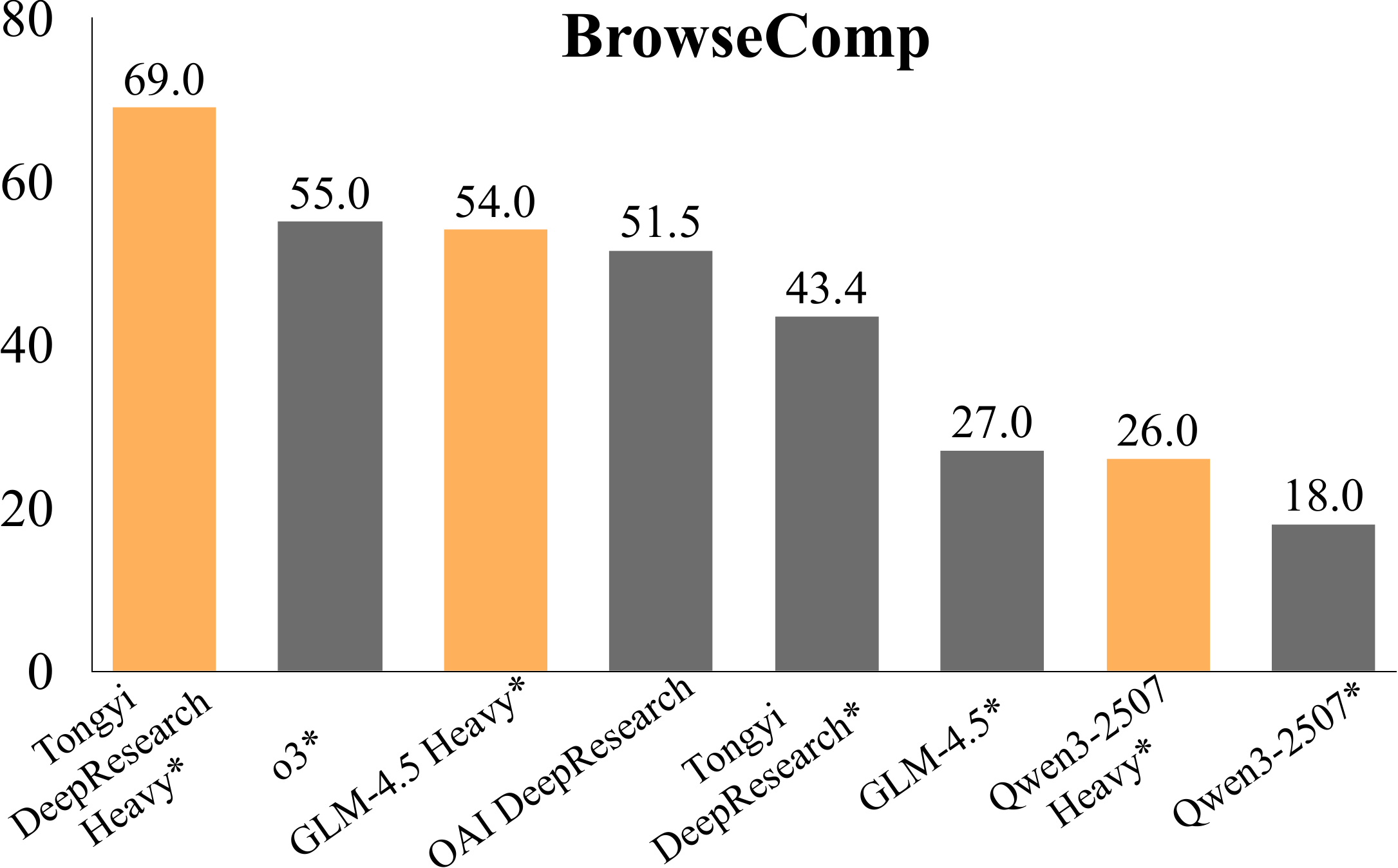}
        }
        \hfill
        \subfigure{
                \includegraphics[width=0.48\textwidth]{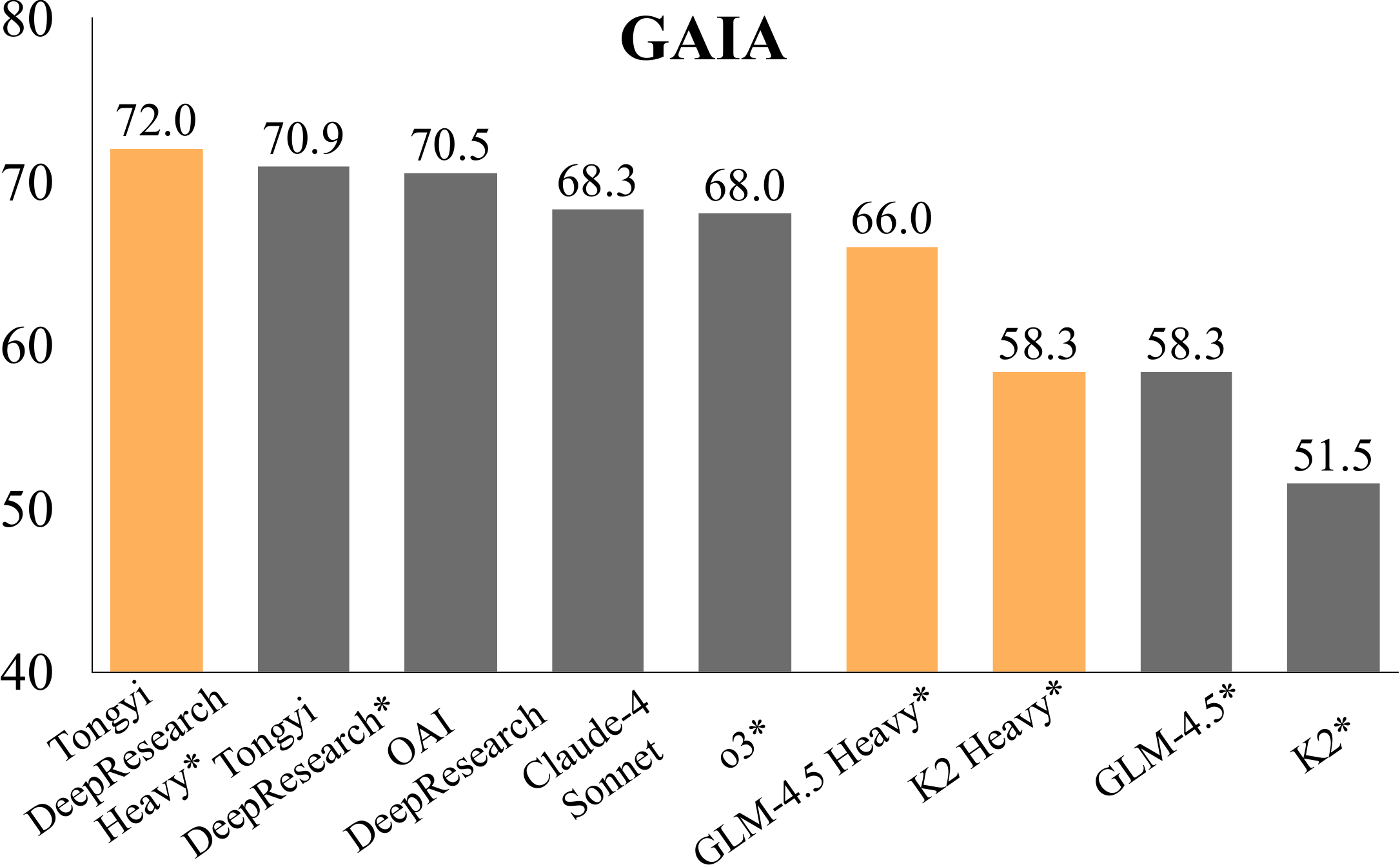}
        }
    \end{minipage}
    \vspace{-10pt}
 \subfigure{  
\includegraphics[width=0.96\textwidth]{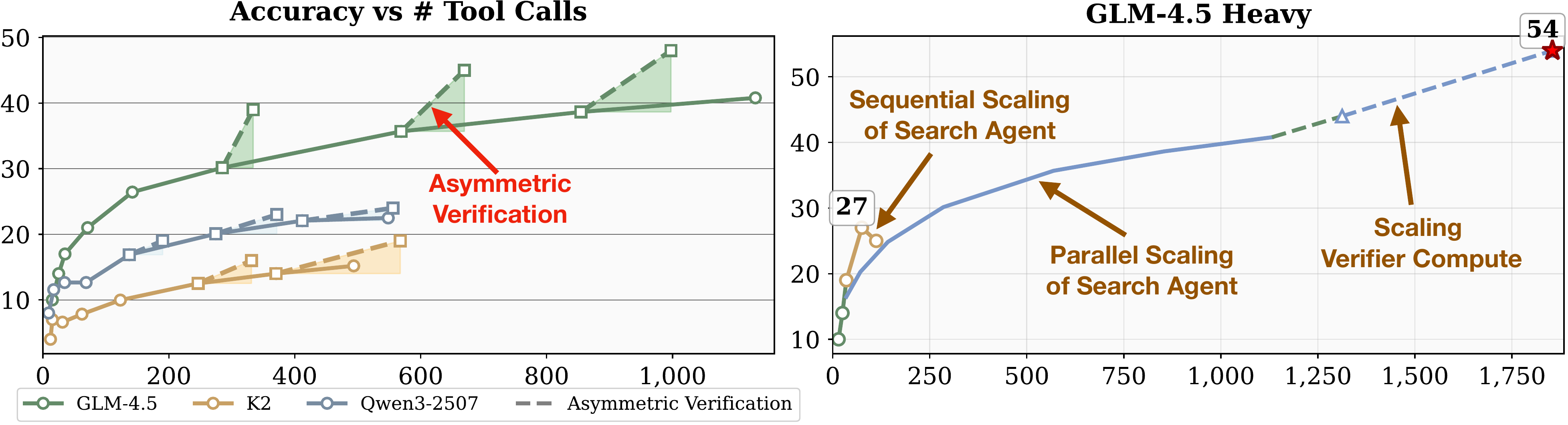}}
\caption{\textbf{Top part} shows accuracy on BrowseComp and GAIA. Results marked with * are from our test runs; \emph{*-Heavy} denotes the accuracy after our test-time scaling. \textbf{Bottom left} shows how accuracy on BrowseComp varies with tool calls. Solid lines indicate scaling the search agent’s compute, dashed lines indicate allocating compute to a verifier. \textbf{Bottom right} shows strategies for extending GLM-4.5 to GLM-4.5 Heavy on BrowseComp.
        }
        \label{fig:figure1}

\end{figure}

\section{Introduction}
Test-time scaling underpins the progress of most advanced AI systems, such as Grok 4 Heavy~\citep{xai2025grok4}, GPT-5-Thinking-Pro~\citep{openai2025gpt5systemcard}, and  Gemini 2.5 Pro~\citep{deepmind2025gemini2_5pro}. It operates through two complementary strategies~\citep{snell2024scaling}: sequential scaling and parallel scaling. Sequential scaling deepens reasoning by extending chains of thought, as in OpenAI-o1~\citep{jaech2024openai}, DeepSeek-R1~\citep{guo2025deepseek}, and Kimi-k1.5~\citep{team2025kimia}, where reinforcement learning encourages reflective reasoning that can unravel complex tasks. Parallel scaling, by contrast, broadens the exploration space by producing multiple candidate outputs and selecting among them through aggregation methods such as Best-of-K selection. This strategy is widely used in advanced closed-source systems such as GPT-5-Thinking-Pro~\citep{openai2025gpt5systemcard} and Gemini-2.5 Pro~\citep{deepmind2025gemini2_5pro}. While expanding exploration space greatly increases the likelihood of capturing the correct solution, it also imposes steep computational costs.




Compared to spending test-time compute on endlessly expanding the number of candidate outputs, allocating compute to verification to evaluate and select outputs introduces a new and orthogonal scaling dimension. Many tasks, such as Sudoku, exhibit an \emph{Asymmetry of Verification}~\citep{wei2025asymmetry}: they are far easier to verify than to solve, meaning that allocating compute to verifiers may yield disproportionate benefits. Many applications of deep search are prime examples of this property. In deep search, the models are often required to search deeply over the web through multiple recursive searching operations, browse hundreds or even thousands of pages, and identify information that satisfies the users' needs.
In such scenarios, forward search must navigate an enormous, sparsely informative space, while backward verification begins with a candidate answer and checks whether it satisfies well-defined conditions~\citep{wei2025browsecomp}, drastically shrinking the search space and making far more efficient use of compute. Research on test-time scaling for deep search, particularly leveraging this asymmetry, therefore opens the path to building powerful and cost-effective agentic search systems such as Deep Research~\citep{openai2025deepresearch,google2025geminideepresearch,xai2025grok3beta}.

In this work, we use the deep search capabilities of Deep Research systems as a representative case to study how scaling test-time compute affects complex information-seeking behavior. Under a simple agentic scaffold, we first investigate strategies that encourage models to increase search tool usage within a single trajectory. For example,
we adopt the \emph{budget forcing} strategy of~\citet{muennighoff2025s1}. By forcing models to make additional tool calls after premature termination, we incentivize exploration of alternative reasoning paths.
This substantially boosts tool usage and Pass@1 accuracy: GLM-4.5~\citep{zeng2025glm} improves from 19\% to 27\%, and Qwen3-0527 from 8\% to 18\% on the challenging BrowseComp benchmark~\citep{wei2025browsecomp}. However, excessive budget forcing eventually degrades performance. To address this, we turn to \emph{parallel scaling}, where expanding compute in parallel further strengthens exploration: GLM-4.5 achieves 67\% Pass@32 accuracy on BrowseComp. However, these gains expose a critical bottleneck: scaling compute broadens the search space but does not help the search agent effectively identify the best candidates—pointing to the need to shift compute from search toward verification.


Deep search benefits from this shift because of the \emph{Asymmetry of Verification}, where verifying an answer may be much easier than generating one.
For instance, on BrowseComp, GLM-4.5 needs about 75 search tool calls on average to solve a problem, but verifying a candidate answer requires only about 18 tool calls. We leverage this asymmetry by adapting the search agent into a verifier agent with minimal modifications, which produces substantial efficiency gains (see Figure~\ref{fig:figure1} bottom left). On BrowseComp, scaling search agent alone is costly: for GLM-4.5, raising accuracy from 30\% to 40\% requires roughly 500 additional tool calls. In contrast, adding a verifier achieves the same 10-point accuracy increase with only about 100 extra calls. Moreover, verifier performance scales well with additional compute, where budget forcing and parallel sampling further improve the accuracy on filtering correct answers.

These results highlight a compute-optimal paradigm for test-time scaling. Rather than allocating all resources to exploration, dedicating a substantial share of compute to verification yields disproportionately larger gains. By striking the right balance between search agents and verifier agents, selecting scaling strategies appropriate to the task (sequential or parallel), and applying effective aggregation metrics, we transform open-source models such as GLM-4.5, K2, Qwen3-2507 and Tongyi-DeepResearch into their ``Heavy'' variants through test-time scaling as exemplified in  Figure~\ref{fig:figure1} bottom right. In this way, open-source models achieve performance levels comparable to leading commercial systems as exemplified in Figure~\ref{fig:figure1} top part. Particularly, GLM-4.5 Heavy, an open-source system, performs on par with commercial leaders such as OpenAI DeepResearch and o3 across most benchmarks, achieving 54.0\% accuracy on BrowseComp, 49.0\% on BrowseComp-zh, 66.0\% on GAIA, and 68.0\% on xbench-DeepSearch. Tongyi-DeepResearch, designed for long-horizon information-seeking, further boosts performance, with the ``Heavy'' variant reaching 69\% accuracy on BrowseComp.



\section{Scaling Search Compute}
\label{sec:scaling_searcher}
In this section, we begin by presenting the framework of the search agent and methods for scaling compute. We then outline the experimental setup, specifying the benchmarks, models, and evaluation metrics employed. Lastly, we report the experimental results on scaling the search agent's test-time compute within the proposed framework.

\subsection{Overview of the Search Agent Framework}
\label{sec:search_agent_framework}
To ensure scalability and generality, we design a streamlined framework based on ReAct~\citep{yao2023react}. In this framework, the search agent handles user problems by iteratively reasoning, generating and executing actions, and processing observations received from real-world web environments. The agent's action space includes either generating a final answer or invoking search tools. The search tool itself combines the web search and browsing functionalities introduced in WebThinker~\citep{li2025webthinker}: it takes as input a search query and an explicit search intent, and returns organized information retrieved from the web. Within this tool, an auxiliary model determines how to organize retrieved content or whether additional browsing is needed. To study test-time compute scaling in a controlled setting, we select K2 as the auxiliary model across the entire paper and do not change it (see Appendix~\ref{sec:impact_aux} for details on this choice). Similar frameworks with searching and browsing functionalities based on ReAct have been widely adopted to develop web searching agents~\citep{li2025websailor,liu2025webexplorer,gao2025beyond}.

\subsection{Test-Time Scaling Approaches}
We categorize methods for scaling the search agent's compute into two types:  

\paragraph{Sequential Scaling} increases computational resources along a single agentic trajectory for each individual problem. We introduce two straightforward yet effective methods: (1) \emph{Max \# Tool Call}: It is common practice to establish a maximum number of tool calls allowed for the search agent, which is explicitly defined in the agent's system prompt. Once the agent reaches this limit, no additional results are provided from the search tool. We experiment with this approach to encourage more tool callings, and details of the system prompt configuration are available in Figure~\ref{fig:searcher-prompt}. (2) \emph{Budget Forcing}: Our experiments revealed that agents often terminate tasks prematurely, even when the system prompt permits sufficient tool calls. Inspired by the approach in~\citet{muennighoff2025s1}, after the agent produces an initial final answer, we allocate additional budget for further tool usage and instruct the agent to continue exploring alternative solution paths or search strategies. Examples of budget forcing scenarios can be found in the Figure~\ref{fig:budget_forcing_example}.

\paragraph{Parallel Scaling} expands computation by generating multiple independent solution trajectories for the same problem in parallel and then applying aggregation methods to determine the final output. We classify these aggregation methods into two types, depending on whether they rely on an external verifier: (1) \emph{Majority Voting}: This approach follows the self-consistency principle~\citep{wang2022self}, where the model independently produces multiple trajectories, and the most frequently predicted answer is selected as the final output. However, majority voting has limited practicality, as it cannot easily extend to open-ended generation tasks where answers are diverse and not strictly comparable. (2) \emph{Verifier-Based}: These approaches employ an external verifier to assign scores to the sampled trajectories. Based on these scores, one can either adopt \emph{Best-of-K}, which selects the trajectory with the highest score, or \emph{Weighted Voting}, which aggregates candidate trajectories by weighting them according to their verifier-assigned scores. In what follows, we first report results obtained using \emph{Majority Voting} where we focus on analyzing the search agent alone, while in \textsection\ref{sec:scaling_verifier} we present experiments incorporating verifier-based strategies.

\begin{figure}[t]
\vspace{-10pt}
        \centering
\includegraphics[width=0.98\textwidth]{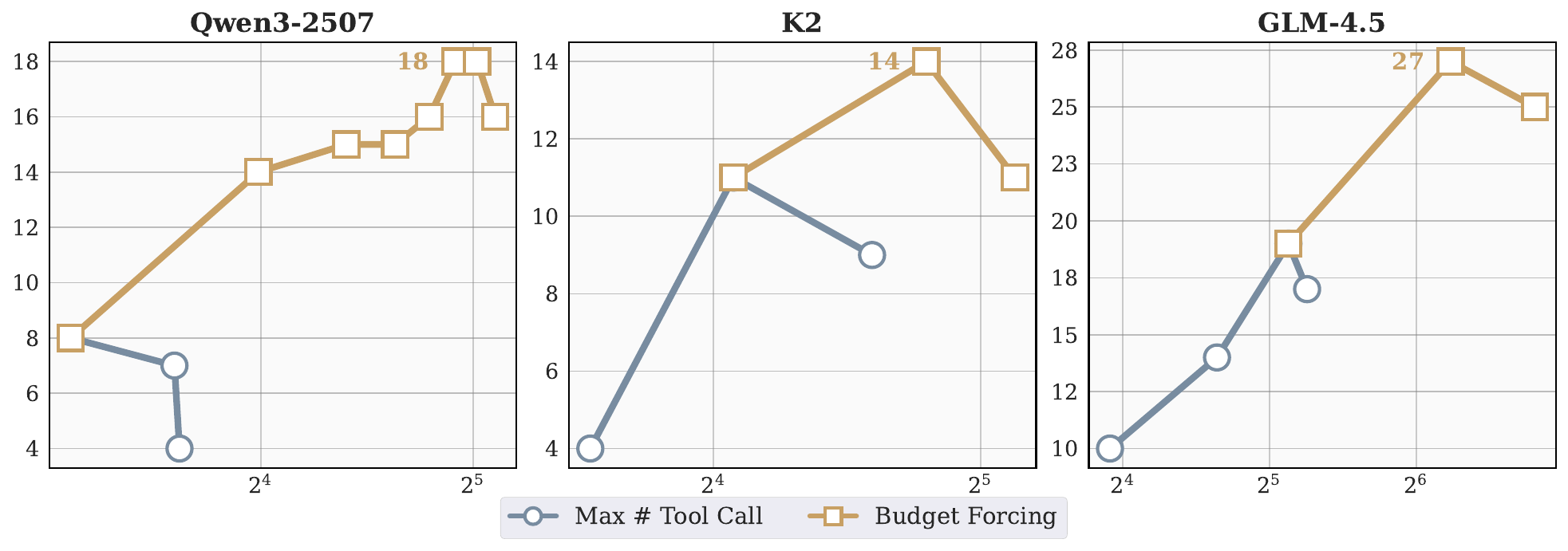}
\vspace{-10pt}
\caption{Sequential scaling of compute, with the x-axis representing the actual number of tool calls, and the y-axis representing Pass@1 accuracy on BrowseComp. For Max \# Tool Call, search tool limits are set to 15, 30, and 50 for Qwen3-2507 and K2, and 15, 30, 50, and 100 for GLM-4.5. Budget Forcing begins from each model’s peak Max \# Tool Call setting and expands until saturation: Qwen3-2507 starts at 15, adding 15 tools per step (7 expansions); K2 starts at 30, adding 30 per step (2 expansions); GLM-4.5 starts at 50, adding 50 per step (2 expansions).
        }
        \label{fig:scaling_policy_seq}
    \vspace{-15pt}

\end{figure}


\subsection{Experimental Setup}
\label{sec:scaling_search_setup}
\paragraph{Benchmarks and Models}

To rigorously evaluate the agent’s information-seeking capabilities, we use benchmarks that are substantially more challenging than standard QA tasks. For this section and \S\ref{sec:scaling_verifier}, we use BrowseComp~\citep{wei2025browsecomp} to study test-time scaling in a series of controlled experiments, where the search queries are complex and require navigation through a vast space of potential answers. Since the complete BrowseComp includes about 1200 questions, completing the full test consumes substantial computational resources. Therefore, a random sample of 100 is used to reduce costs while still being representative enough to reflect performance on the full dataset. In \S\ref{sec:heavy_version}, we will include more benchmarks to present the final test-time scaling results. We select three open-source models with strong reasoning and agentic capabilities, GLM-4.5~\citep{zeng2025glm}, K2~\citep{team2025kimi}, and Qwen3-235B-A22B-Instruct-2507~\citep{yang2025qwen3}, as search agents to evaluate their effectiveness when scaling test-time compute.

\paragraph{Metrics}
Following prior work~\citep{gao2025beyond,li2025websailor}, we approximate the test-time compute of search agents by counting the actual number of tool calls they make to solve each problem. This choice reflects the nature of deep research tasks, which often require interacting with external tools to gather information. 
To evaluate the performance of the search agent, we use two metrics: \textbf{Pass@K}, which measures whether the search agent finds the correct answer at least once among K independently sampled trajectories; and \textbf{Maj@K}, which measures whether the correct answer matches the majority vote among the answers from K independently sampled trajectories. 
\subsection{Experimental Results}

\paragraph{The Max \# Tool Call method underutilizes compute, while Budget Forcing boosts tool use and performance.} Figure~\ref{fig:scaling_policy_seq} illustrates how different models scale with test-time compute on BrowseComp under two methods: Max \# Tool Call and Budget Forcing. Under the Max \# Tool Call setting, even when models are allotted generous tool call quotas, they often terminate early. For instance, Qwen3-2507 shows diminishing returns—increasing its quota from 15 to 50 results in only marginal increases in tool usage and even performance drops. In contrast, Budget Forcing actively drives models to exploit available compute. After a single application of budget forcing, GLM-4.5's tool usage more than doubles, and its Pass@1 accuracy rises from 19\% to 27\%. Qwen3-2507 exhibits an even more pronounced jump, with its performance ceiling increasing from 8\% to 18\% across multiple rounds. However, we observe diminishing gains as the number of tool calls within a single trajectory increases. This saturation suggests that long trajectories challenge models to perform coherent long-range reasoning.

\begin{figure}[t]
\vspace{-10pt}
        \centering
\includegraphics[width=0.98\textwidth]{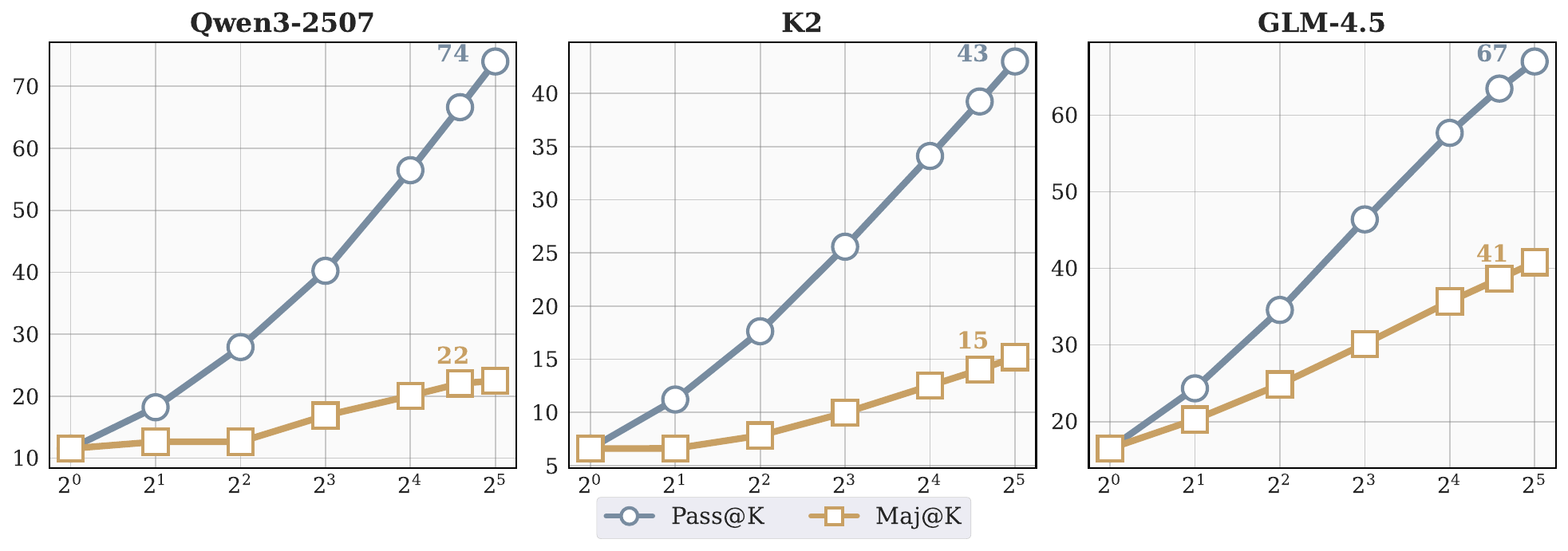}
\vspace{-10pt}
\caption{Parallel scaling of compute, where the x-axis shows K and the y-axis shows Pass@K and Maj@K accuracy on BrowseComp. For K2 and GLM-4.5, we first apply the Max \# Tool Call strategy, setting maximum tool usage to 30 and 50 respectively, then perform parallel scaling by independently sampling K = 1, 2, 4, 8, 16, 24, 32 trajectories. For Qwen3-2507, we apply the Max \# Tool Call strategy with a maximum of 15 tools, then apply Budget Forcing to add 15 more, followed by parallel scaling with K = 1, 2, 4, 8, 16, 24, 32 trajectories.
        }
        \label{fig:scaling_policy_pal}
    \vspace{-15pt}

\end{figure}

\paragraph{Models exhibit strong exploration capabilities but struggle with exploitation.} As shown in Figure~\ref{fig:scaling_policy_pal}, surprisingly, increasing compute via parallel scaling leads to a rapid rise in Pass@K accuracy, indicating that with more sampled trajectories, models are increasingly likely to find at least one correct answer. For instance, GLM-4.5's Pass@K accuracy increases from 16\% to 67\% when scaling K from 1 to 32 on the extremely challenging BrowseComp benchmark. 
Figure~\ref{fig:combine_budget_parallel} in Appendix~\ref{sec:budget_forcing_parallel} further shows that combining Budget Forcing with parallel scaling enhances exploration even more. However, this does not translate into efficient exploitation: while K2 reaches a Pass@16 of about 34\%, its Maj@16 accuracy—reflecting the ability to identify the correct answer through majority voting—is only around 12\%. This gap suggests that simply encouraging broader exploration is insufficient. Instead, improving the model’s ability to recognize and select high-quality answers is critical. This suggests that instead of simply promoting more diverse exploration, we should focus on improving the model's ability to exploit promising candidates. In the following sections, we explore how incorporating external verification signals can improve exploitation efficiency in \textsection\ref{sec:scaling_verifier} and how balancing exploration with exploitation~\citep{zeng2024b,liu2024diving} leads to better overall performance in \textsection\ref{sec:heavy_version} .

\section{Scaling Verification Compute}
\label{sec:scaling_verifier}
In this section, we begin by introducing the \emph{asymmetric verification}, followed by the framework of the verifier we implemented. We then outline the experimental setup and present the results.


\subsection{Asymmetric Verification}

Asymmetric Verification has become a central concept in AI, highlighting tasks where verifying a solution is far easier than generating one. Many problems have this ``easy to verify'' property. For instance, in Sudoku or the Eight Queens problem, generating a valid solution requires extensive search, but once a candidate solution is given, verification is trivial by simply checking whether all constraints are satisfied. Some tasks, however, show near-symmetry of verification, where checking correctness takes almost as much effort as solving. A classic example is matrix multiplication: confirming the product of two large matrices is essentially as costly as performing the multiplication itself. At the other extreme, certain problems are hard to verify: writing code may be relatively straightforward, but rigorously verifying that it is free from security vulnerabilities can demand exhaustive formal verification or full security audits. This ``easy to verify'' property allows the construction of robust reward systems at relatively low cost. By leveraging inexpensive verification, AI performance can be significantly boosted through scaling during training or at test time.

The deep search task similarly exhibits this asymmetry. While forward search requires models to navigate a vast space and retrieve hard-to-find information—often through extensive exploration—verification only needs minimal effort to confirm whether a predicted answer satisfies the necessary conditions. Consider the following example from BrowseComp:

\begin{tcolorbox}[colback=blue!5!white, colframe=black]
Please provide the name of a video game released in 1992. The series' first game was featured in a globally recognized record book, and it earned a spot in an Australian photographer publication. The third game inspired a legacy sequel. The fifth installment was the first to have a Chinese dub.
\end{tcolorbox}
Given a video game, one can easily verify its validity via simple web searches by checking its release year, whether the first game was in the record book, and so on. We also evaluate different models on search and verification tasks across multiple benchmarks, measuring the number of tool calls each requires in Table~\ref{table:tool_call_number}. The results show that most deep research tasks exhibit asymmetric verification, and this asymmetry grows with task difficulty. For instance, in BrowseComp, GLM-4.5 uses about 75.3 tool calls to search for a candidate answer but only 18 tool calls to verify it. In this section, we aim to harness this asymmetry by introducing verifiers to filter candidate answers, thereby improving the efficiency of test-time compute usage.

\subsection{Overview of The Verifier Framework}
Our verifier shares the same design principles, framework, and search tools as the search agent in \textsection\ref{sec:search_agent_framework}, differing only in the system prompt. In the verifier’s prompt, the model is instructed to use search tools specifically to check the correctness of a predicted answer and to assign a confidence score based on whether the answer meets the necessary conditions confirmed by the search results. The full system prompt is provided in Figure~\ref{fig:verfier-prompt} in the Appendix.

\subsection{Experimental Setup}

We use the same open-source models introduced in \S\ref{sec:scaling_search_setup} and evaluate them on the most challenging benchmark, BrowseComp. To fairly compare the computational efficiency of scaling search agents versus scaling verifier agents, we start from the models that have already undergone parallel scaling. From this baseline, we explore two directions: (1) further scaling the search agent's computation, continuing to generate more trajectories through parallel sampling (2) introducing a verifier agent to evaluate the predicted answers from the existing trajectories. For the verifier-based approach, we filter candidates using the verifier's confidence scores, testing strategies such as Best-of-K and Weighted Voting. We note that verifier computation can also be scaled sequentially (e.g., Budget Forcing) or in parallel, where multiple verification trajectories are generated for each candidate answer. The final confidence score is the average across trajectories, reducing bias from any single sample.

\begin{table*}[t]
\centering

\caption{Average number of tool calls for searching and verification across models and benchmarks. Since xbench-DeepSearch has relatively simple search requirements and balanced difficulty between search and verification, we report only the tool calls needed for search.}
\vspace{-5pt}
\resizebox{0.9\textwidth}{!}{
\begin{tabular}{lccccccc}
\toprule
\textbf{Model} & \multicolumn{2}{c}{\textbf{BrowseComp}} 
               & \multicolumn{2}{c}{\textbf{BrowseComp-zh}} 
               & \multicolumn{2}{c}{\textbf{GAIA}} 
               & \textbf{xbench-DeepSearch} \\ 
\cmidrule(lr){2-3} \cmidrule(lr){4-5} \cmidrule(lr){6-7} \cmidrule(lr){8-8}
& \textbf{Solve} & \textbf{Verify} 
& \textbf{Solve} & \textbf{Verify} 
& \textbf{Solve} & \textbf{Verify} 
& \textbf{Solve} \\ 
\midrule
Qwen3-2507 & 32.4 & 11.3 & 18.8 & 9.7 & 9.7 & 7.9 & 13.4 \\
K2        & 27.8 & 11.2 & 38.2 & 6.3 & 6.2 & 5.8 & 2.9  \\
GLM-4.5    & 75.3 & 18.0 & 48.3 & 9.0 & 17.4 & 10.6 & 6.1 \\
\bottomrule
\end{tabular}

}
\label{table:tool_call_number}
\vspace{-15pt}
\end{table*}

\subsection{Experimental Results}




\begin{figure}[!t]
        \centering
\includegraphics[width=0.97\columnwidth]{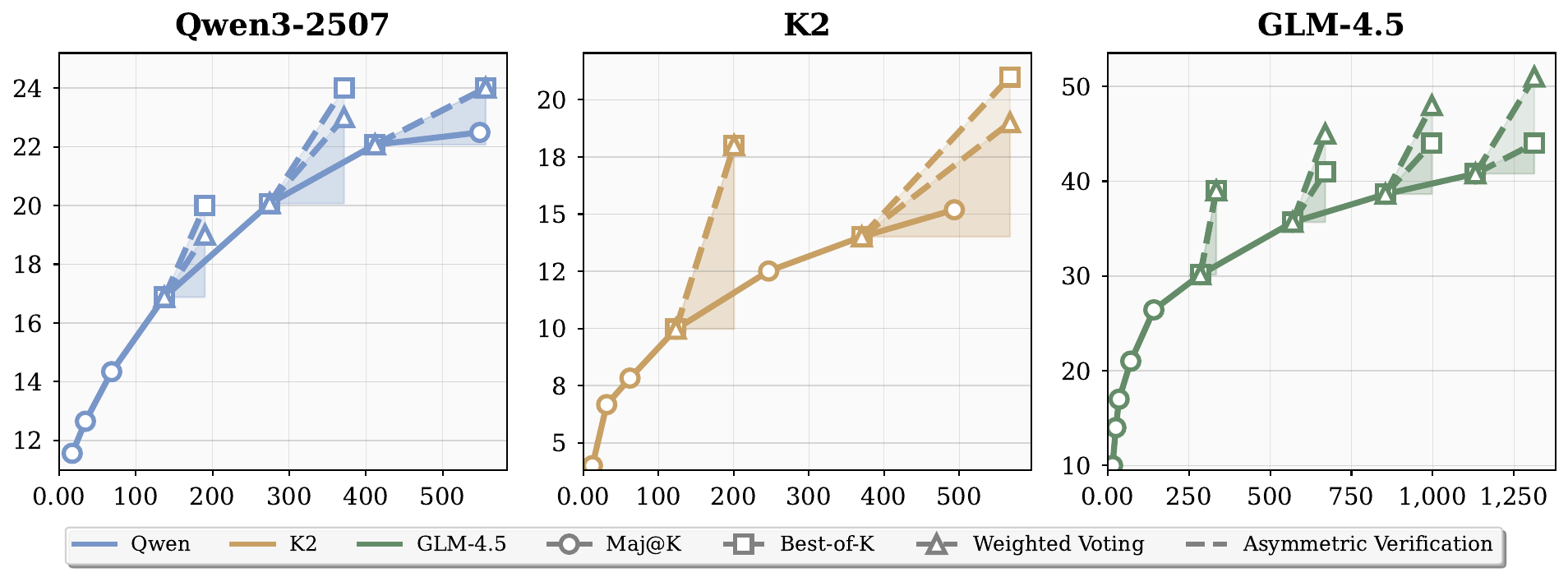}
 \vspace{-5pt}
\caption {Parallel scaling results of different models on BrowseComp.  The x-axis shows the number of tool calls counting both the searcher and the verifier. The solid lines represent the growth of Maj@K corresponding to scaling the searcher's compute, while the dashed lines represent the growth of Best-of-K and Weighted Voting after introducing a verifier.} 
        \label{fig:verfier_advantage}
\end{figure}

\paragraph{Verifier yields superior accuracy–cost trade-off.} 
Figure~\ref{fig:verfier_advantage} illustrates that the asymmetry property of verification consistently appears across different models when scaling test compute. In other words, adding a verifier allows accuracy to grow more quickly than simply increasing the computation of the search agent. For example, with GLM-4.5, increasing search computation alone requires about an additional 560 tool calls just to raise accuracy from 35.7\% with Maj@8 to 40.8\% with Maj@32. By contrast, introducing a verifier achieves a larger gain with far less cost: only about 100 additional tool calls are needed to improve from 35.7\% with Maj@8 to 45\% with Weighted Voting. A similar pattern emerges with K2. Without a verifier, raising performance from around 10\% with Maj@8 to 15.2\% with Maj@32 requires an increase of about 370 tool calls. With a verifier, however, the same starting point improves to 18\% with Best-of-8 or Weighted Voting after only about 77 additional tool calls.



\begin{figure}[!t]
        \centering
\includegraphics[width=0.97\columnwidth]{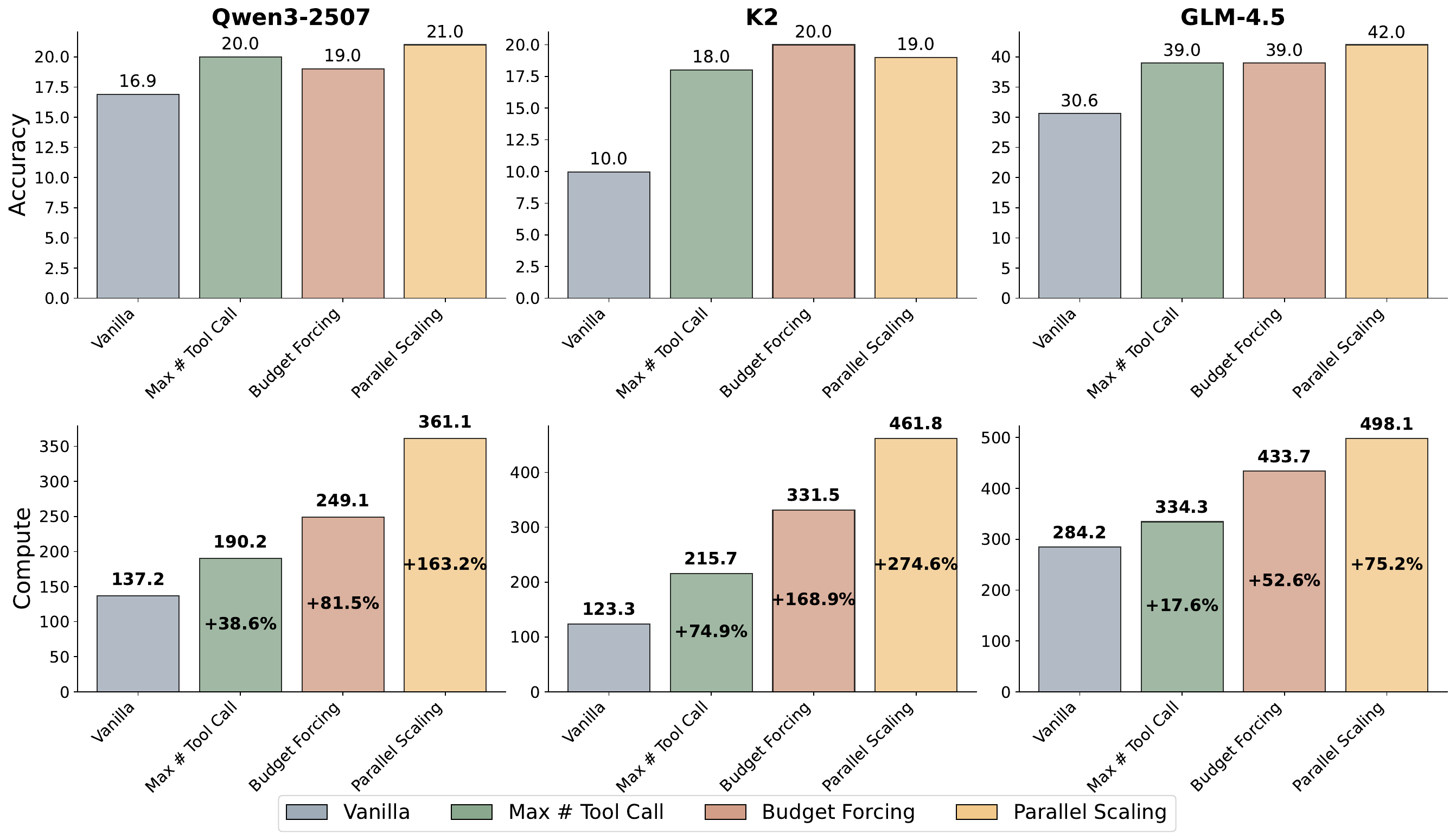}
\caption {Different strategies for scaling verifier computation across models. The top panel shows accuracy, and the bottom panel shows the corresponding number of tool calls. Along the x-axis, vanilla indicates the Maj@8 accuracy achieved by scaling the search agent without verification, while Max \# Tool Call, Budget Forcing, and Parallel Scaling show the Best-of-8 results when applying these strategies to increase verifier compute.}
        \label{fig:policy_verifier_scaling_comparison}
\vspace{-15pt}
\end{figure}

\paragraph{Scaling the verifier can further raise the performance ceiling, though the gains depend on the model and strategy used.} We apply the scaling methods like  Max \# Tool Call, Budget Forcing and Parallel Scaling to the verifier, with results shown in Figure~\ref{fig:policy_verifier_scaling_comparison}. These methods lead to notable improvements. For example, in the K2 case, applying Budget Forcing increases actual tool usage and boosts accuracy from 10\% to 20\%.
Scaling methods such as budget forcing or parallel scaling, and answer aggregation schemes such as Best-of-N or Weighted Voting produce distinct improvement patterns. For example, with GLM-4.5, parallel scaling paired with the Best-of-8 approach achieves the largest gain, reaching 42\% on BrowseComp. In contrast, K2 attains better Best-of-8 accuracy with lower compute using budget forcing. These results show that, in practical deployments, the choice of scaling target (e.g., search agent or verifier agent), scaling strategy, and answer aggregation metric should be guided by both the model’s characteristics and the available compute budget, to maximize performance per unit of cost. 


\section{Pushing Test-Time Scaling Limits of Open Models}
\label{sec:heavy_version}

\begin{table*}[t]
\centering
\caption{Accuracy (\%) of proprietary and open-source agentic systems across multiple benchmarks. Results marked with \textdagger{} are from our test runs, indicating the maximum Pass@1 that models can achieve when provided with sufficient tool calls without budget forcing. For Tongyi DeepResearch, we report both the results of our test runs and the officially reported results}
\resizebox{0.9\textwidth}{!}{
\begin{tabular}{lcccc}
\toprule
                    & \textbf{BrowseComp}  & \textbf{BrowseComp-zh} & \textbf{GAIA} & \textbf{xbench-DeepSearch} \\
\midrule
\multicolumn{5}{c}{Proprietary Deep-Research Systems / Models} \\
\midrule
Gemini 2.5 Pro & 7.6 & 27.3 & - & - \\
OpenAI o1 & 9.9 & 29.1 & - & - \\
OpenAI o3\textsuperscript{\textdagger}         & 55.0  & 59.0 & 68.0 & 68.0 \\
OpenAI GPT-5 & 19.8 & 34.3 & - & 30.0 \\
Grok3 DeepResearch  &       & 12.9 & -    & 50+ \\
Doubao DeepResearch & -     & 26.0 & -    & 50+ \\
OpenAI DeepResearch & 51.5 & 42.9 & 70.5 & 66.7 \\
\midrule
\multicolumn{5}{c}{Open-Source Models} \\
\midrule
DeepSeek-R1 & 2.0 & 23.2 & 16.5 & 32.7 \\
Qwen3-2507\textsuperscript{\textdagger}          & 8.0  & 23.0 & 44.7 & 45.5 \\
K2\textsuperscript{\textdagger}                 & 11.0 & 22.0 & 50.0 & 54.0 \\
GLM-4.5\textsuperscript{\textdagger}             & 19.0 & 27.0 & 58.0 & 58.0 \\
DeepSeek-V3.1 & 30.0 & 49.2 & - & 71.2 \\
Tongyi DeepResearch\textsuperscript{\textdagger} & 43.0 & 39.0 & 70.6 & 68.0 \\
Tongyi DeepResearch & 43.4 & 46.7 & 70.9 & 75.0 \\
\midrule
\multicolumn{5}{c}{Open-Source Models (Heavy Version)} \\
\midrule
\rowcolor{gray!12} Qwen3-2507 Heavy
& \textbf{29.0} \textcolor{green!50!black}{(+21.0)}
& \textbf{42.0} \textcolor{green!50!black}{(+19.0)}
& \textbf{53.4} \textcolor{green!50!black}{(+8.7)}
& \textbf{63.0} \textcolor{green!50!black}{(+17.5)} \\
\rowcolor{gray!12} K2 Heavy
& \textbf{24.0} \textcolor{green!50!black}{(+13.0)}
& \textbf{36.0} \textcolor{green!50!black}{(+14.0)}
& \textbf{58.3} \textcolor{green!50!black}{(+8.3)}
& \textbf{57.0} \textcolor{green!50!black}{(+3.0)} \\
\rowcolor{gray!12} GLM-4.5 Heavy
& \textbf{54.0} \textcolor{green!50!black}{(+35.0)}
& \textbf{49.0} \textcolor{green!50!black}{(+22.0)}
& \textbf{66.0} \textcolor{green!50!black}{(+8.0)}
& \textbf{68.0} \textcolor{green!50!black}{(+10.0)} \\
\rowcolor{gray!12} Tongyi DeepResearch Heavy
& \textbf{69.0} \textcolor{green!50!black}{(+26.0)}
& \textbf{55.0} \textcolor{green!50!black}{(+16.0)}
& \textbf{72.8} \textcolor{green!50!black}{(+2.2)}
& \textbf{80.0} \textcolor{green!50!black}{(+12.0)} \\
\bottomrule
\end{tabular}
}
\vspace{-5pt}
\label{table:result_with_tts}
\vspace{-15pt}
\end{table*}

\subsection{Experimental Setup}

Building on the results from \S\ref{sec:scaling_searcher} and \S\ref{sec:scaling_verifier}, we can conceptualize test-time compute scaling through three orthogonal dimensions: scaling target, scaling strategy, and aggregation metric. \textbf{Scaling target} defines what part of the system is being expanded—either the search agent, which explores candidate solutions, or the verifier agent, which evaluates and filters them.  
\textbf{Scaling strategy} specifies how the additional compute is applied, encompassing approaches such as Max \# Tool Call  (adjusting the maximum number of tool calls allowed), Budget Forcing (actively increasing tool invocation), and Parallel Scaling (running multiple trajectories concurrently).
\textbf{Aggregation metric} determines how the outputs are combined and assessed, using criteria like Pass@1 (single-best answer), Maj@K (majority voting), Weighted Voting (confidence-weighted voting), or Best-of-K (selecting the highest-scoring candidate). This unified view helps disentangle where to allocate extra computation, how to deploy it, and how to measure its benefits, clarifying trade-offs and guiding compute-optimal scaling decisions. 

\paragraph{Benchmarks and Models}
Compared to evaluating the agent on QA tasks that are relatively simple, we aim to assess its information-seeking capabilities using significantly more challenging benchmarks. Specifically, we evaluate on the following benchmarks: (1) BrowseComp~\citep{wei2025browsecomp}, which includes complex search problems that require navigating through a large space of potential answers, where relevant information is difficult to locate. (2) BrowseComp-zh~\citep{zhou2025browsecomp}, the Chinese counterpart of BrowseComp to evaluate LLM agents on the Chinese web. (3) GAIA~\citep{mialon2023gaia}, which evaluates multi-modality and tool-use capabilities; following previous work~\citep{li2025websailor}, we focus on its text-only validation set. (4) xbench-DeepSearch~\citep{chen2025xbench}, another challenging benchmark designed to assess tool usage capabilities specifically in search and information retrieval contexts. The models used remain consistent with those in \S\ref{sec:scaling_searcher} and \S\ref{sec:scaling_verifier}. We also include Tongyi-DeepResearch-30B-A3B~\citep{tongyidr}, a model that is specifically designed for long-horizon, deep information-seeking tasks. Our evaluation uses 100 randomly sampled tasks from BrowseComp and BrowseComp-zh, since running the full test is computationally expensive. This sampling balances cost efficiency with representativeness. In contrast, evaluations for other benchmarks are conducted on their full datasets.

\paragraph{Scaling Details} 

We scale performance by combining Max \# Tool Call scaling with Budget Forcing to reach the Pass@1 peaks of different models. To broaden the exploration space of the search agents, we apply parallel scaling on top of sequential scaling. For example, on BrowseComp, GLM-4.5 generates 32 parallel samples, K2 generates 24, and Qwen3-2507 generates 16, while on other benchmarks every model samples  8 times in parallel. Once these trajectories are produced, a verifier selects the answers, after which parallel scaling or budget forcing is applied again to the verifier to increase the verifier’s test-time compute. 
These scaling approaches lead to the heavey variant of different models.
We document the specific scaling setup for GLM-4.5 Heavy, K2 Heavy, and Qwen3-2507 Heavy in Appendix~\ref{sec:scaling_details_heavy_variant}.


\subsection{Experimental Results}
\begin{figure}[!t]
        \centering
\includegraphics[width=0.99\columnwidth]{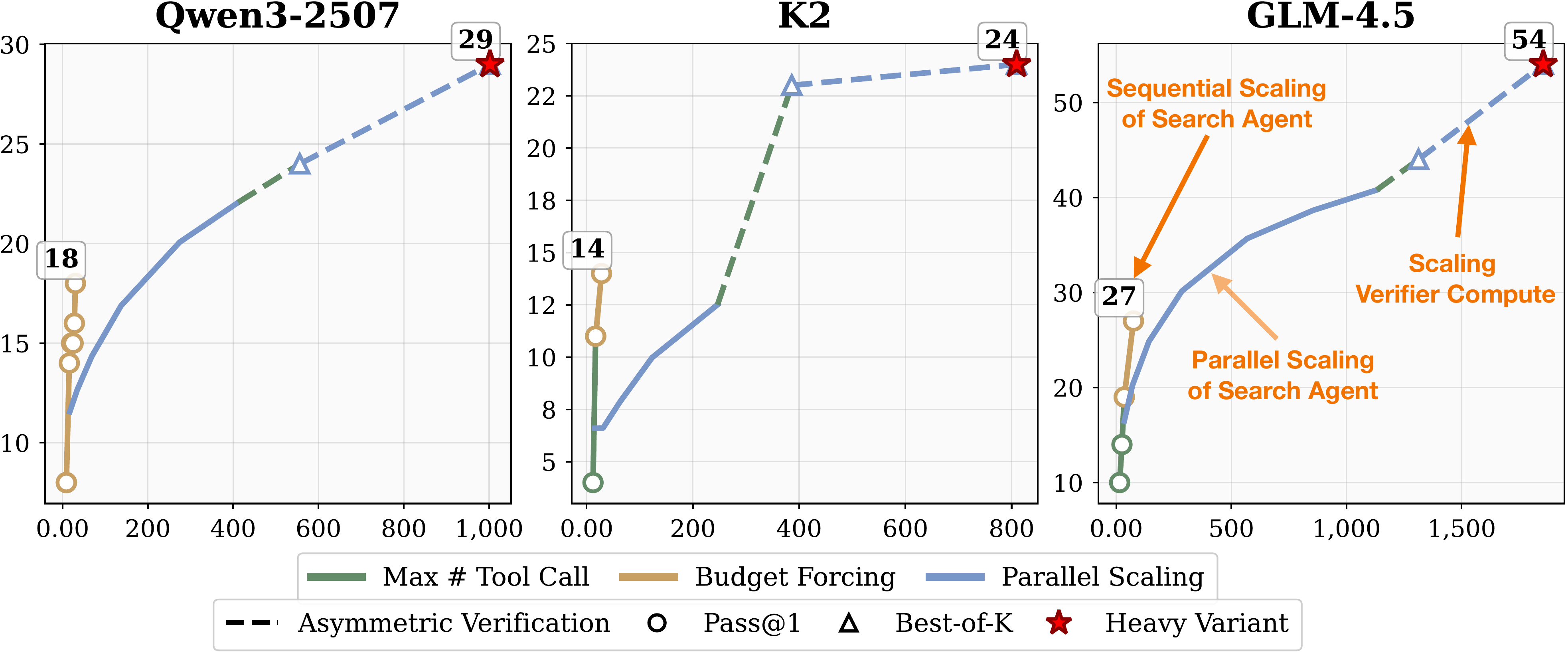}
\vspace{-10pt}
\caption{Scaling test time compute for different models on BrowseComp, where the x-axis represents the actual number of tool calls, and the y-axis represents accuracy.}
        \label{fig:browse_experiment_results_comparisonv2}
\vspace{-15pt}
\end{figure}

Table~\ref{table:result_with_tts} presents the experimental results of GLM-4.5 Heavy, K2 Heavy, and Qwen3-2507 Heavy. The three models after test-time scaling achieve substantial performance gains across all benchmarks. For example, Qwen3-2507 Heavy improves accuracy over Qwen3-2507 by about 20 absolute points on BrowseComp, BrowseComp-zh, and xbench-DeepSearch. Notably, GLM-4.5 Heavy reaches 54.0\% accuracy on BrowseComp, 66.0\% on GAIA, and 68.0\% on xbench-DeepSearch, putting it on par with the best commercial agents such as o3 and OpenAI Deep Research on these three benchmarks. We also introduce a heavy variant of Tongyi Deep Research, which achieves 69\% accuracy on BrowseComp, 55\% accuracy on BrowseComp-zh, 72.8\% on GAIA and 80.0\% on xbench-DeepSearch as shown in Figure~\ref{fig:figure1}. Full implementation details are provided in the Appendix~\ref{sec:scaling_details_heavy_variant}.

Figure~\ref{fig:browse_experiment_results_comparisonv2} presents the results of scaling test-time compute for different models on BrowseComp. Sustained compute expansion can bring open-source systems close to closed-source performance. Verifier introduction further accelerates performance gains. For K2, adding a verifier increases Best-of-16 to 23\%, with a slope much steeper than simply expanding the search agent. Figure~\ref{fig:browse-zh_experiment_results_comparisonv2} in Appendix~\ref{sec:scaling_browsecomp_zh} shows that performance patterns on BrowseComp-zh closely mirror those on BrowseComp. On BrowseComp-zh, GLM-4.5 attains a Best-of-8 score of 49\%, exceeding the level of OpenAI Deep Research (see Table \ref{table:result_with_tts}). Notably, parallel scaling of the verifier yields consistent accuracy gains: for K2, Best-of-8 rises from 31\% to 36\%. This is because parallel sampling mitigates variance in individual confidence scores, making the selection process more robust to noise by reducing the impact of extreme values. Figure~\ref{fig:gaia_experiment_results_comparisonv2} in Appendix ~\ref{sec:scaling_gaia} presents the results of scaling test-time compute for different models on GAIA. Across models, different scaling strategies still yield notable performance gains for both search agents and verifier agents. We report xbench-DeepSearch results in Figure~\ref{fig:xbench_experiment_results_comparisonv1} of the Appendix. Because solving and verifying tasks in this benchmark are nearly equally difficult (as detailed in Appendix~\ref{sec:scaling_xbench}), we only evaluate search-agent scaling in Figure~\ref{fig:xbench_experiment_results_comparisonv1} and Table~\ref{table:result_with_tts}.

\section{Discussion}
In this paper, we present a systematic study of test-time scaling for deep research agents, examining both sequential and parallel scaling. Our approach primarily exploits the properties of asymmetric verification, where we apply test-time scaling to a range of flagship open-source models, extending them into ``heavy variants'' that achieve substantially higher performance on deep research tasks. Although our current use of asymmetric verification is relatively simple by employing a verifier to select final answers, it nevertheless proves highly effective, significantly improving the efficiency of test-time scaling. Looking ahead, we aim to adopt more flexible strategies. For example, verifiers could be applied at each step along the search trajectory, or even used to actively guide the search process. In addition, we envision incorporating asymmetric verification more deeply into training, enabling search agents to internalize verification capabilities and using them directly during inference.



\bibliography{iclr2026_conference}
\bibliographystyle{iclr2026_conference}
\newpage
\appendix


\section{The Use of Large Language Models}
Large language models (LLMs) were used exclusively to refine the writing of this paper. All aspects of the research process—including ideation, design, analysis, and interpretation—were conducted entirely without the assistance of LLMs.

\section{Impact of Auxiliary Model on Search Tool}
\label{sec:impact_aux}

\begin{figure}[!t]
        \centering
\includegraphics[width=0.5\columnwidth]{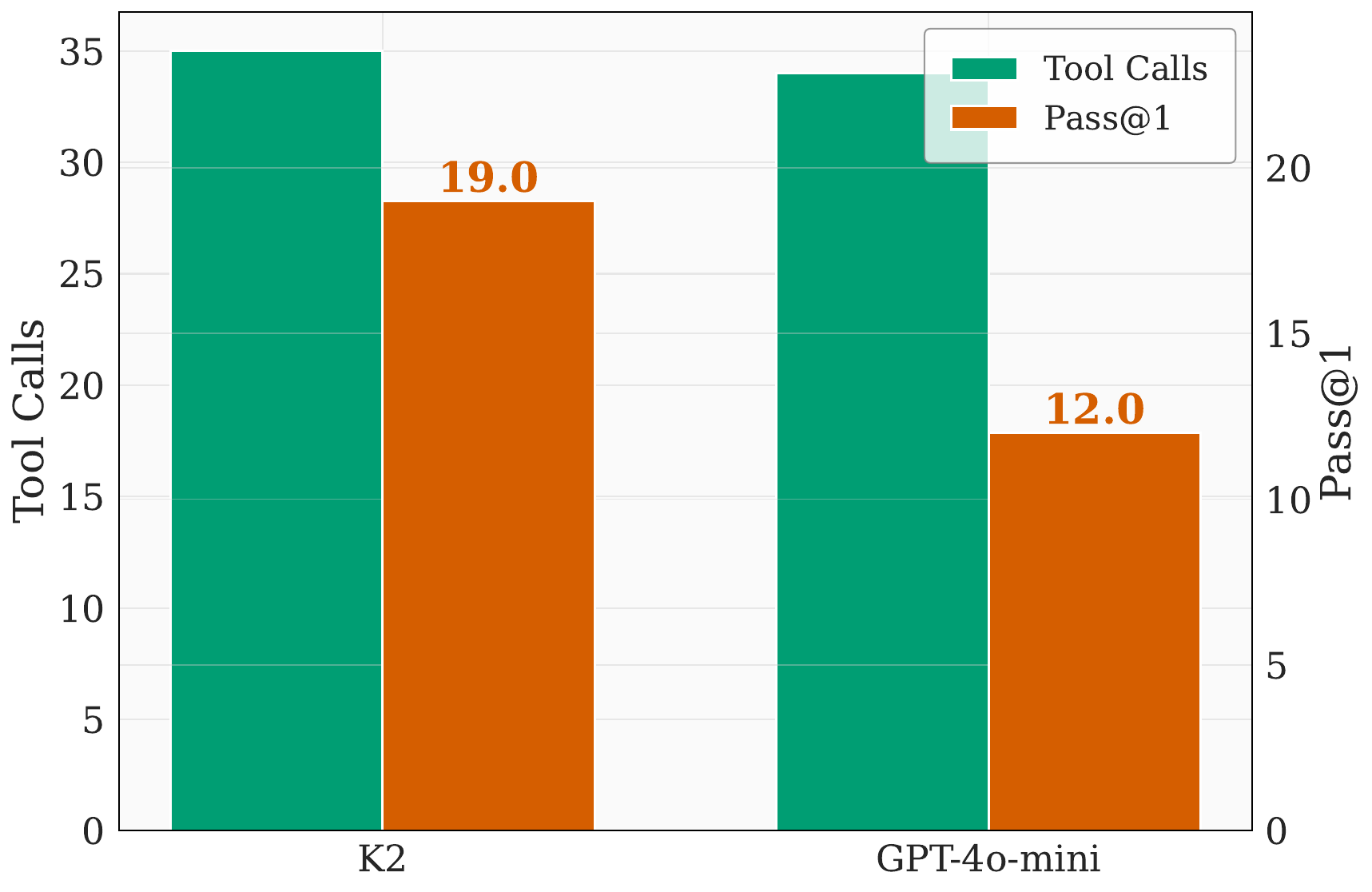}
\vspace{-10pt}
\caption{Pass@1 and the average number of actual tool uses per problem for different auxiliary models, under a maximum limit of 50 search tool calls.}
        \label{fig:aux_model_comparison}
        \vspace{-10pt}
\end{figure}

The auxiliary model in the search tool is responsible for extracting information from retrieved content based on the search intent, initiating follow-up queries, and selectively accessing relevant URLs. We assess how its capabilities affect search tool performance by using GLM-4.5 as the search agent and comparing a stronger auxiliary model (K2) with a weaker one (GPT-4o-mini) with a maximum of 50 tool calls. Figure~\ref{fig:aux_model_comparison} shows performance on BrowseComp along with the actual number of tool calls. While both auxiliary models make a similar number of calls, their accuracy diverges sharply: GPT-4o-mini reaches only 12\%, whereas K2 attains 19\%. This gap demonstrates that auxiliary model strength directly impacts the quality of search results and, consequently, the search agent’s final performance. For this reason, we use K2 as the default auxiliary model in our experiments.


\section{Combining Budget Forcing with Parallel Scaling}
\label{sec:budget_forcing_parallel}
\begin{figure}[!t]
        \centering
\includegraphics[width=0.7\columnwidth]{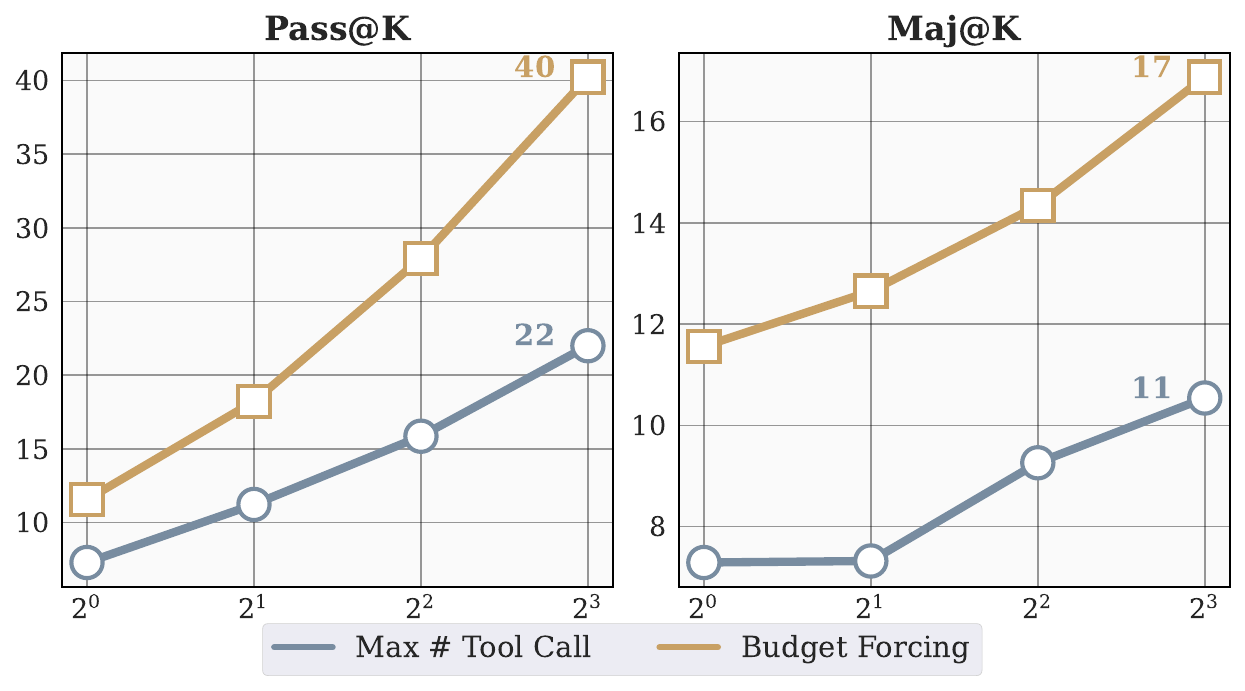}
\vspace{-10pt}
\caption {A comparison of Qwen3-2507's performance after applying Max \# Tool Call and Budget Forcing followed by parallel scaling. The x-axis represents K; the y-axis shows Pass@K and Maj@K. Max \# Tool Call limits tool usage to 15 calls and applies Parallel Scaling with K = 1, 2, 4, 8 independent trajectories. Budget Forcing increases this limit by 15 calls and uses the same K values.} 
        \label{fig:combine_budget_parallel}
        \vspace{-10pt}
\end{figure}

In Figure~\ref{fig:combine_budget_parallel}, we compare Qwen3-2507's performance after applying Max \# Tool Call and Budget Forcing, followed by parallel scaling. The results show that combining Budget Forcing with parallel scaling further enhances exploration.


\section{Scaling Details of the Heavy Variant}
\label{sec:scaling_details_heavy_variant}

We scale performance by combining Max \# Tool Call scaling with Budget Forcing to reach the Pass@1 peaks of different models. In BrowseComp, Budget Forcing is applied to Qwen3-2507 by expanding it five times, with each expansion adding 15 additional tool calls. For K2, the expansion is performed once with 30 additional tool calls, while GLM-4.5 is expanded once with 50 additional tool calls. In BrowseComp-zh, Budget Forcing is also applied to Qwen3-2507, this time expanding it three times with 30 additional tool calls at each step. K2 achieves its highest Pass@1 performance when the Max \# Tool Call is set to 15, whereas GLM-4.5 is expanded once with 50 additional tool calls. In GAIA, Budget Forcing is again used to expand Qwen3-2507, adding 15 additional tool calls at each of three expansion steps. K2 is expanded once with 30 additional tool calls, while GLM-4.5 achieves its peak Pass@1 score when the Max \# Tool Call is set to 30. Finally, in xbench-DeepSearch, Qwen3-2507 is expanded five times, with each step adding 15 additional tool calls. K2 reaches its highest Pass@1 score when the maximum number of tool calls is set to 100, whereas GLM-4.5 reaches its peak with the maximum number of tool calls set to 15.

To broaden the exploration space of the search agents, we apply parallel scaling on top of sequential scaling. On BrowseComp, for instance, GLM-4.5 generates 32 parallel samples, K2 generates 16, and Qwen3-2507 generates 24, while on other benchmarks every model produces 8 parallel samples. Once these trajectories are generated, a verifier evaluates the answers, after which we apply either parallel scaling or budget forcing again to the verifier in order to increase its test-time compute. On BrowseComp with Qwen3-2507, each trajectory from the search agent is verified four times in parallel, and the verifier’s average score is taken as the trajectory score, yielding a Best-of-24 accuracy. With K2, the same process produces a Best-of-16 accuracy, and with GLM-4.5 it yields a Best-of-32 accuracy. On BrowseComp-zh, the procedure is similar: Qwen3-2507, K2, and GLM-4.5 all rely on four parallel verifications per trajectory, with the average score determining the outcome, leading in each case to a Best-of-8 accuracy. On GAIA, Qwen3-2507 and K2 follow the same approach and also obtain Best-of-8 accuracy. For GLM-4.5 on GAIA, however, we used budget forcing to expand the verifier’s compute once, with the resulting verification score taken as the trajectory score, again producing a Best-of-8 result. Finally, since solving and verifying tasks in xbench-DeepSearch are of nearly equal difficulty, we did not employ a verifier on that benchmark.

We apply test-time extension to Tongyi Deep Research on BrowseComp. Because this model does not perform well as a verifier, we rely on GLM 4.5 for verification. The process works as follows: Tongyi Deep Research first generates 16 parallel samples, and GLM 4.5 then extracts answers from these responses. Next, GLM 4.5 serves as a verifier to check the extracted answers. Finally, through weighted voting, this approach achieves an accuracy of 69\%. For other benchmarks such as BrowseComp-zh, GAIA and xbench-DeepSearch, we take similar steps, where Tongyi Deep Research first generates 8 parallel samples, and GLM 4.5 then extracts answers from these responses. Next, GLM 4.5 serves as a verifier to check the extracted answers. Finally, through weighted voting, this approach achieves 55.0\%, 72.8\%, and 80.0\% on BrowseComp-zh, GAIA and xbench-DeepSearch respectively.
 
\section{Scaling Results on Browsecomp-zh}
\label{sec:scaling_browsecomp_zh}
\begin{figure}[!t]
        \centering
\includegraphics[width=0.99\columnwidth]{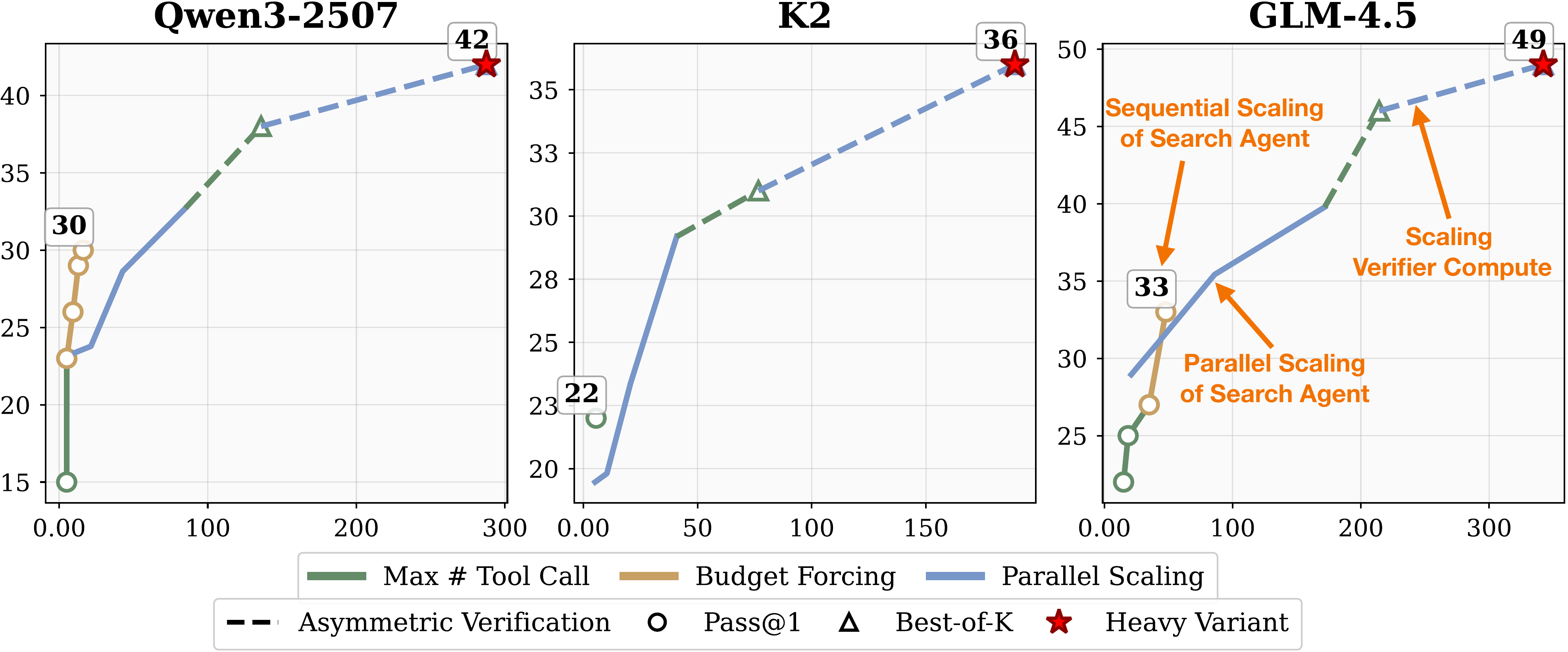}
\vspace{-10pt}
\caption{Scaling test time compute for different models on Browsecomp-zh, where the x-axis represents the actual number of tool calls, and the y-axis represents accuracy.}
        \label{fig:browse-zh_experiment_results_comparisonv2}
        \vspace{-10pt}
\end{figure}

Figure~\ref{fig:browse-zh_experiment_results_comparisonv2} presents the results of scaling test-time compute for various models on Browsecomp-zh.

\section{Scaling Results on GAIA}
\label{sec:scaling_gaia}
\begin{figure}[!t]
        \centering
\includegraphics[width=0.99\columnwidth]{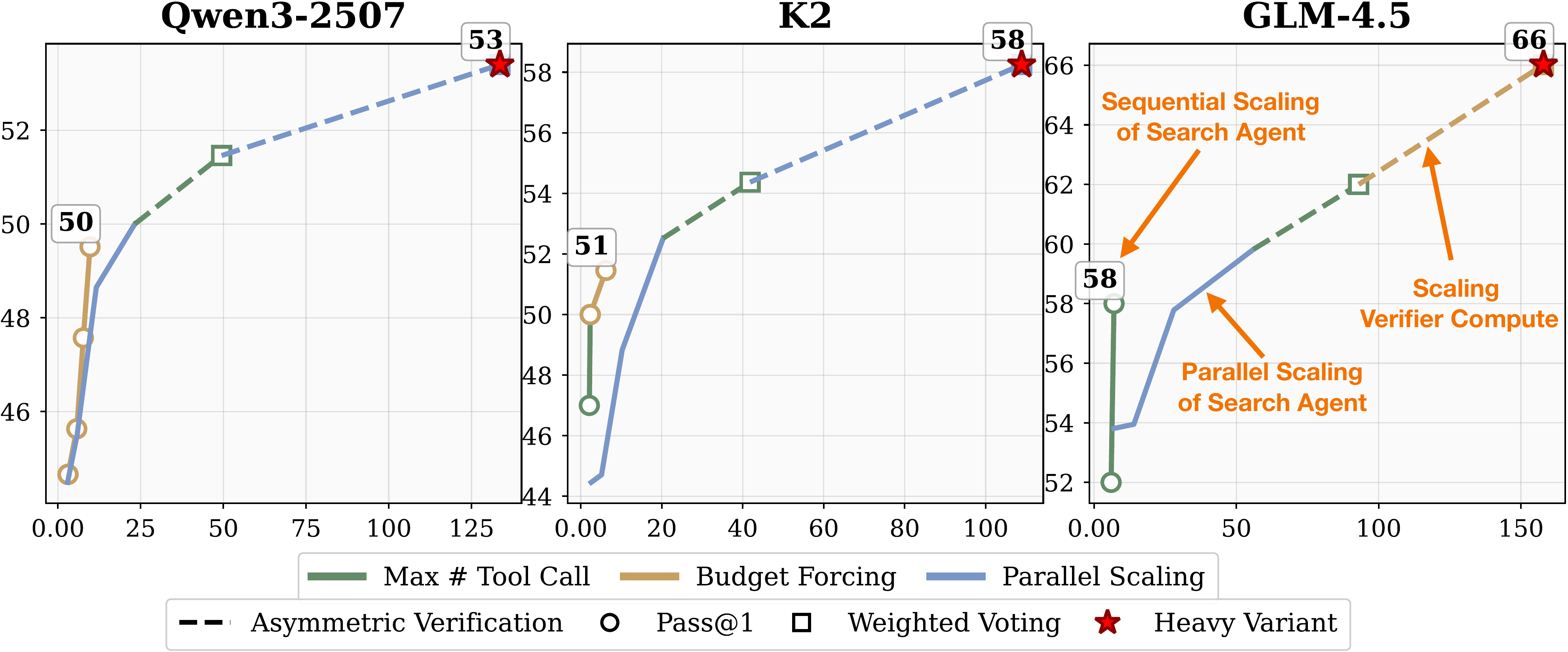}
\vspace{-10pt}
\caption{Scaling test time compute for different models on GAIA, where the x-axis represents the actual number of tool calls, and the y-axis represents accuracy.}
        \label{fig:gaia_experiment_results_comparisonv2}
        \vspace{-10pt}
\end{figure}

Figure~\ref{fig:gaia_experiment_results_comparisonv2} presents the results of scaling test-time compute for various models on GAIA.

\section{Scaling Results on xbench-DeepSearch}
\label{sec:scaling_xbench}
\begin{figure}[!t]
        \centering
\includegraphics[width=0.99\columnwidth]{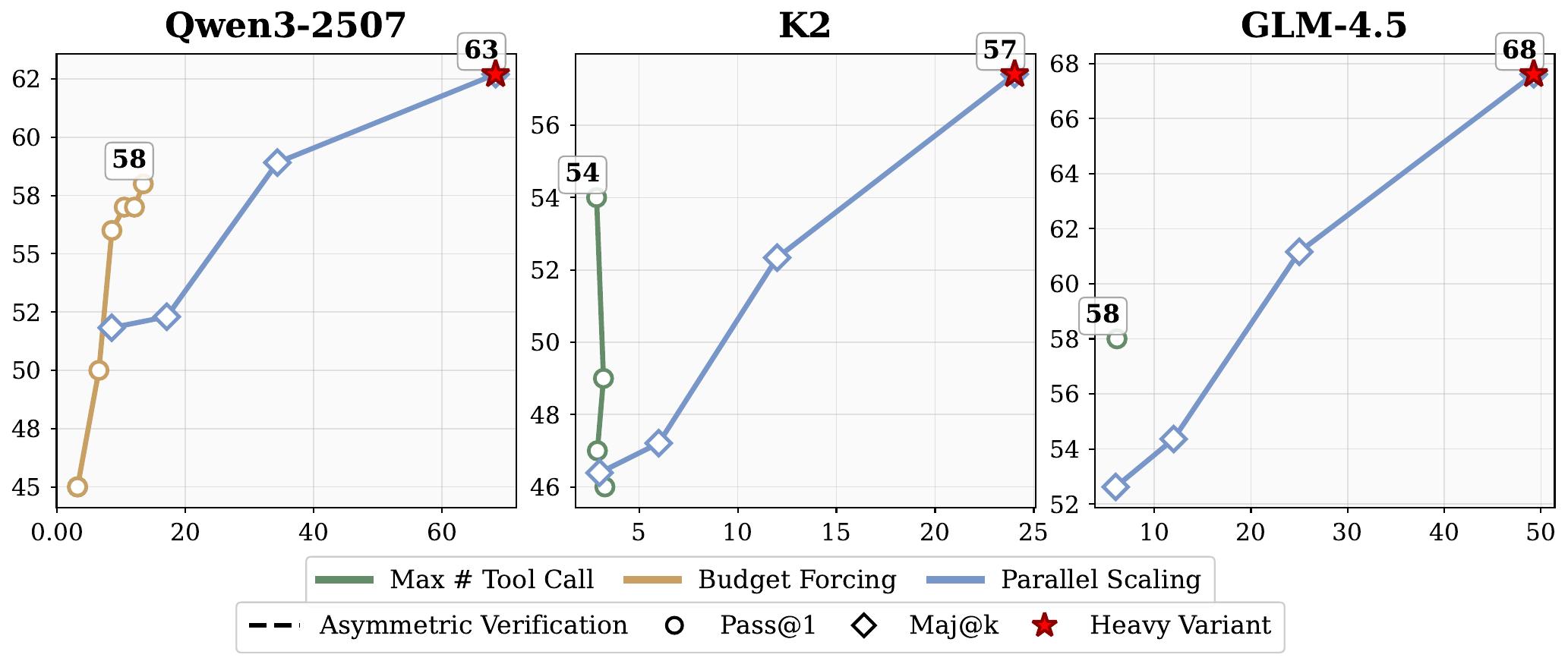}
\vspace{-10pt}
\caption{Scaling test time compute for different models on xbench-DeepSearch, where the x-axis represents the actual number of tool calls, and the y-axis represents accuracy.}
        \label{fig:xbench_experiment_results_comparisonv1}
        \vspace{-10pt}
\end{figure}

Figure~\ref{fig:xbench_experiment_results_comparisonv1} presents the results of scaling test-time compute for various models on xbench-DeepSearch. Scaling the search agent’s compute consistently yields substantial performance gains. In contrast, verifier scaling offers little improvement. For example, with GLM-4.5, Maj@8 reaches 67.61\%, yet adding a verifier only raises weighted voting to 68.0\%. This minimal gain arises because, on xbench-DeepSearch, solving a problem and verifying it require comparable effort. For instance, consider the query: \emph{As of May 18 2025, what is the maximum context token count supported by OpenAI’s codex-1 agent?} The correct answer is 192k, but confirming this demands nearly the same search intensity as finding it.

\begin{figure}[!t]
        \centering
\includegraphics[width=0.98\columnwidth]{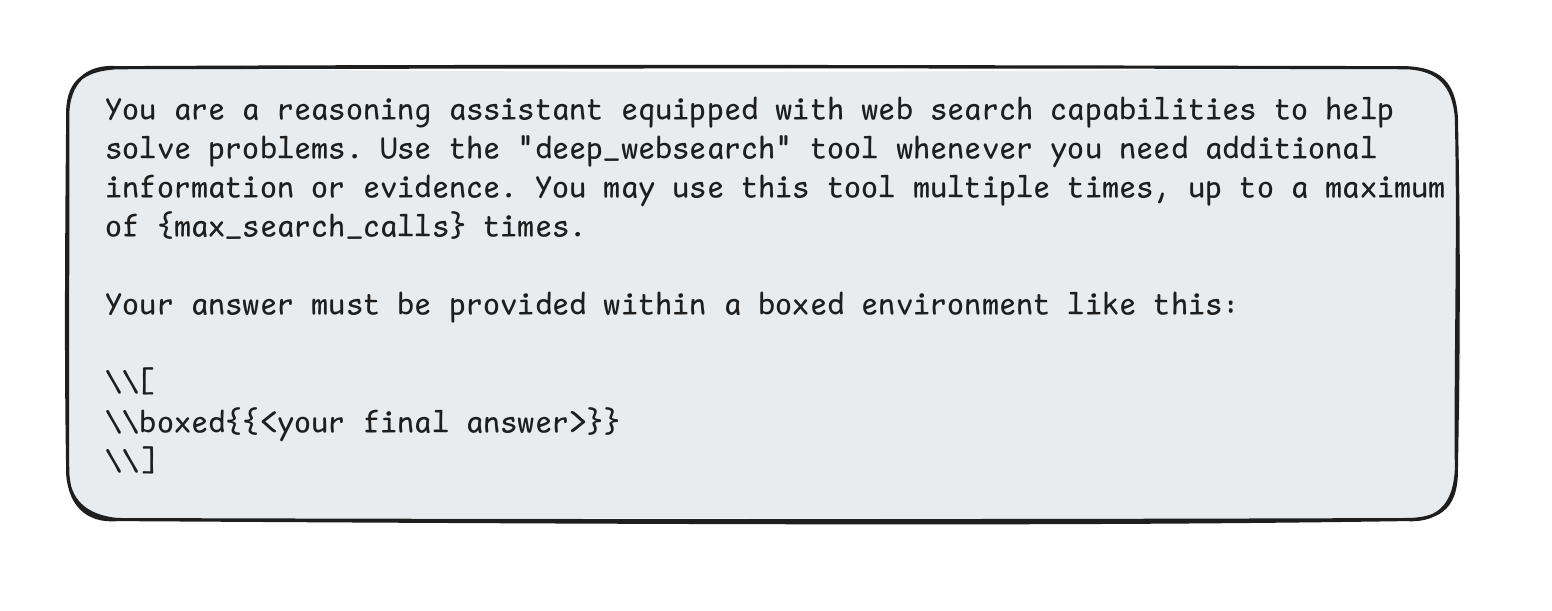}
\vspace{-10pt}
\caption{Details of the Max \# Tool Call's system prompt for search agent.}
        \label{fig:searcher-prompt}
        \vspace{-10pt}
\end{figure}

\begin{figure}[!t]
        \centering
\includegraphics[width=0.98\columnwidth]{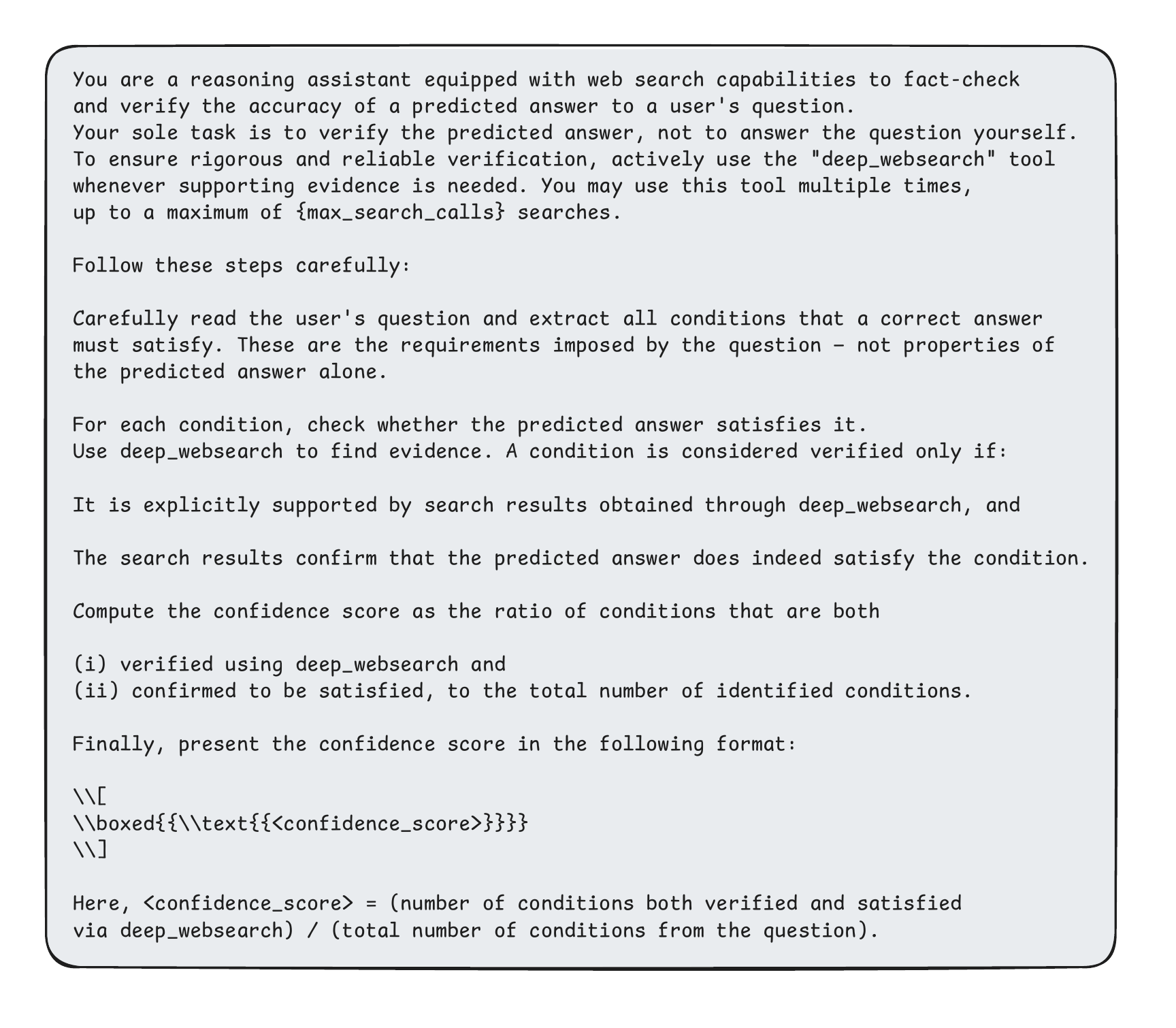}
\vspace{-10pt}
\caption{Details of the Max \# Tool Call's system prompt for verfier agent.}
        \label{fig:verfier-prompt}
        \vspace{-10pt}
\end{figure}

\begin{figure}[!t]
        \centering
\includegraphics[width=0.98\columnwidth]{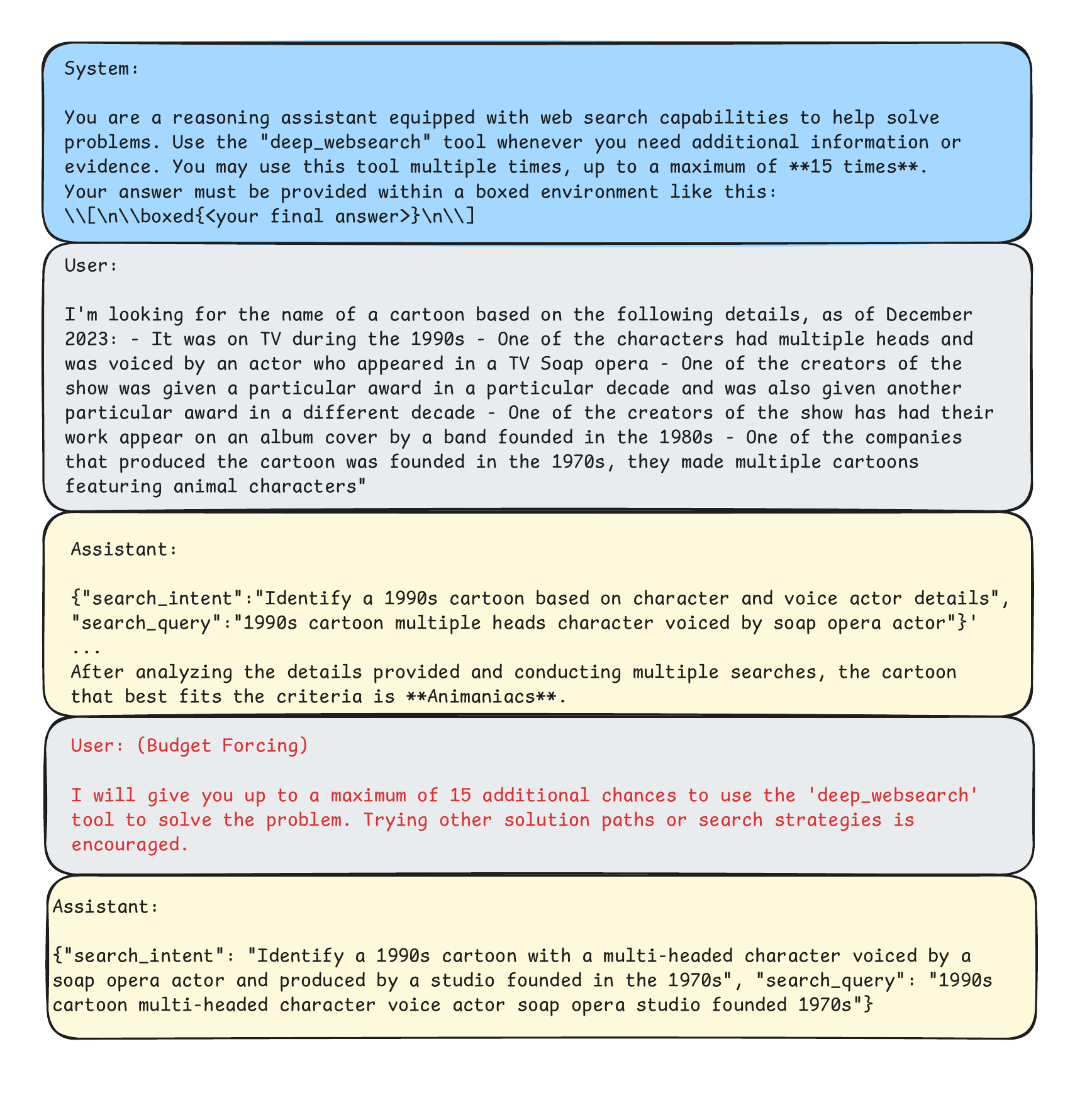}
\caption{Examples of budget forcing scenarios.}
        \label{fig:budget_forcing_example}
\end{figure}


\end{document}